%% file: main.tex
\documentclass[10pt,twocolumn,letterpaper]{article}

\usepackage[pagenumbers]{cvpr} 

\input{preamble}
\usepackage{gensymb}  
\usepackage{textcomp}  
\usepackage{algpseudocode}
\usepackage{algorithm}
\definecolor{cvprblue}{rgb}{0.21,0.49,0.74}
\usepackage[pagebackref,breaklinks,colorlinks,allcolors=cvprblue]{hyperref}

\usepackage{booktabs}       
\usepackage{amsfonts}       

\usepackage{pifont}
\usepackage{multirow}  
\usepackage[table]{xcolor}
\newcommand{\cmark}{\ding{51}} 
\newcommand{\xmark}{\ding{55}} 

\def\paperID{15015} 
\def\confName{CVPR}
\def\confYear{2026}

\title{ArticFlow: Generative Simulation of Articulated Mechanisms}

\author{%
  Jiong Lin
  \quad Jinchen Ruan
  \quad Hod Lipson
  \\
  Creative Machines Lab, Columbia University\\
  New York, NY 10027\\
}

\begin{document}
\maketitle
\input{sec/0_abstract}    
\input{sec/1_intro}

\input{sec/2_related}
\input{sec/3_method}

\input{sec/4_experiments}
\input{sec/5_discussion}
{
    \small
    \bibliographystyle{ieeenat_fullname}
    \bibliography{main}
}

\input{sec/X_suppl}

\end{document}

%% file: preamble.tex
\usepackage{booktabs}
\usepackage{multirow}
\usepackage{graphicx}
\usepackage{array}
\usepackage{gensymb}  
\usepackage{fancyvrb}
\usepackage{lipsum}
\usepackage{xspace}
\usepackage{pifont}
\usepackage{amssymb}
\usepackage{amsmath}
\usepackage{algorithmicx}



%% file: sec/0_abstract.tex
\begin{abstract}
Recent advances in generative models have produced strong results for static 3D shapes, whereas articulated 3D generation remains challenging due to action-dependent deformations and limited datasets. We introduce ArticFlow, a two-stage flow matching framework that learns a controllable velocity field from noise to target point sets under explicit action control. ArticFlow couples (i) a latent flow that transports noise to a shape-prior code and (ii) a point flow that transports points conditioned on the action and the shape prior, enabling a single model to represent diverse articulated categories and generalize across actions. On MuJoCo Menagerie, ArticFlow functions both as a generative model and as a neural simulator: it predicts action-conditioned kinematics from a compact prior and synthesizes novel morphologies via latent interpolation. Compared with object-specific simulators and an action-conditioned variant of static point-cloud generators, ArticFlow achieves higher kinematic accuracy and better shape quality. Results show that action-conditioned flow matching is a practical route to controllable and high-quality articulated mechanism generation.
\end{abstract}

\begin{figure}[t!]
    \includegraphics[width=\columnwidth]{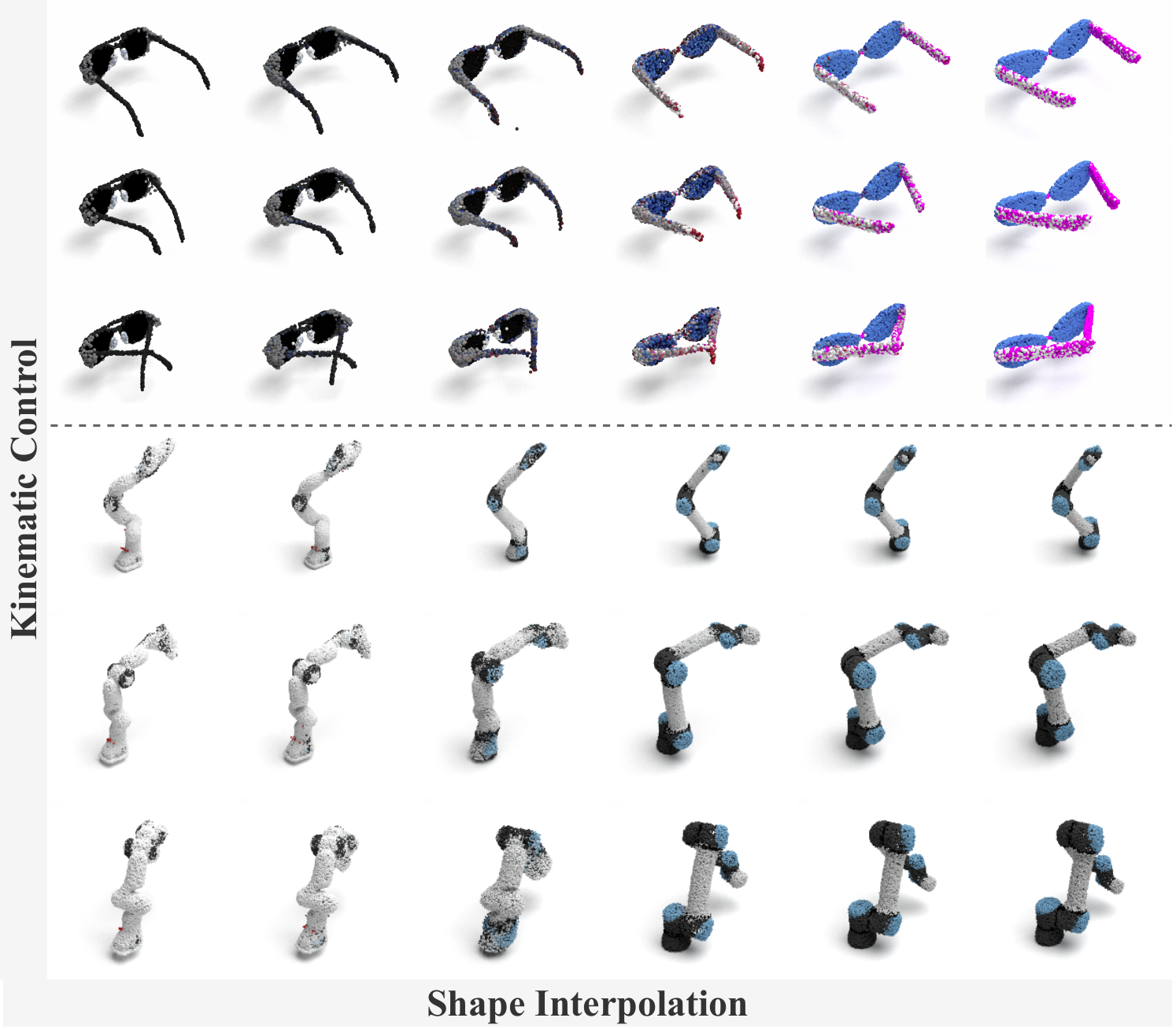}
    \caption{
    \textbf{ArticFlow} models a category of diverse articulated objects with a single point-flow field, conditioned on shape prior and joint actions. In this figure, columns interpolate the shape-prior condition (novel instances); rows sweep the action control (joint angles). Examples include eyeglasses from \textit{PartNet-Mobility}~\cite{xiang2020sapien} and robot arms from \textit{MuJoCo Menagerie}~\cite{tassa2018deepmind}.
    The same field produces coherent deformations for one category of articulated shapes, enabling interpolation and controllable 3D generation.
    }
    \label{fig:1-teaser}

\end{figure}


%% file: sec/1_intro.tex
\section{Introduction}
\label{sec:intro}
Generative models have become a leading approach for content generation across images, audio, and video. The model landscape has progressed from early variational autoencoders (VAEs)~\cite{kingma2013auto} and generative adversarial networks (GANs)~\cite{goodfellow2014generative} to flow based methods such as continuous normalizing flows (CNFs)~\cite{rezende2015variational,papamakarios2021normalizing}, diffusion models~\cite{ho2020denoising, song2020denoising}, and, more recently, flow matching methods~\cite{lipman2022flow}. Building on this progress, recent work has increasingly turned to 3D content~\cite{deitke2023objaverse, deitke2023objaverse-xl}, where geometric structures and physical constraints introduce additional challenges.

Broadly, two competing paradigms have emerged for 3D generation. The first leverages powerful 2D priors to synthesize multi-view consistent images and then lifts them to 3D; for example, Zero-1-to-3~\cite{liu2023zero, shi2023zero123++} style 3D synthesis pipelines. This family benefits from 2D foundation models~\cite{rombach2022high} pre-trained on high-quality image datasets, but often struggles with true 3D consistency and geometric fidelity. They typically rely on a separate reconstruction stage~\cite{mildenhall2021nerf} to convert images into watertight shapes with accurate collisions and physical properties. The second paradigm performs explicit 3D generation, directly producing point clouds~\cite{yang2019pointflow, zhou20213d, wu2023fast}, meshes, or voxels. It depends on high-quality, curated 3D datasets~\cite{chang2015shapenet, mo2019partnet} and typically demands model designs~\cite{qi2017pointnet, qi2017pointnet++} that align with representation-specific properties (e.g., permutation invariance for point clouds, connectivity constraints for meshes, and spatial sparsity for voxels). However, many real-world objects are articulated, coupling geometry with kinematics; this setting stresses both paradigms and motivates dedicated generative models for articulated shapes.

Within this articulated setting, interest is rising, but generative methods remain in their early stages. Most prior work focuses on analysis, such as part segmentation, articulation inference, and motion estimation~\cite{li2020category, lin2025autourdf}. The challenge is to ensure 3D consistency, along with kinematic correctness and plausible deformation. Existing approaches either condition on a given image~\cite{liu2024singapo, chen2025freeart3d} and then reconstruct its articulation, or they follow a multi-stage pipeline that first generates a kinematic graph (parts, joints, parameters) and subsequently synthesizes geometry for each part~\cite{liu2024cage}.

Against this backdrop, we propose a self-contained, 3D-explicit solution for articulated mechanism generation. We represent each object as a deformable point cloud and learn an action conditioned flow matching velocity field that captures kinematic motion from sets of deformation samples. We further condition the field on a shape prior, enabling a single model to capture the kinematics of an entire category of articulated shapes. By interpolating in the action and shape-prior space, the model generates novel instances and motions while maintaining 3D consistency and kinematic validity. As shown in Figure~\ref{fig:1-teaser}, we evaluate across multiple categories from the \textit{PartNet-Mobility}~\cite{xiang2020sapien} dataset and on simulated robots from \textit{MuJoCo Menagerie}~\cite{tassa2018deepmind}. Our model exhibits disentangled conditioning: with the \emph{shape prior} held fixed, generated samples preserve appearance while producing kinematically correct deformations; with the \emph{action} held fixed, generated samples maintain a similar pose while varying appearance to yield novel instances.


Beyond visual novelty, these results point to a broader opportunity: the potential to address the long-standing scarcity of diverse robot morphologies.
High quality articulated objects and simulation ready robot models are expensive. For example, \textit{PartNet-Mobility} includes 26 folding chairs, and only 17 robot arms are available in \textit{MuJoCo Menagerie}. Meanwhile, robot behavior datasets have scaled from RT-1 to RT-X~\cite{brohan2022rt,zitkovich2023rt,o2024open}, reaching $10^5$ tasks with millions of trajectories, yet the number of distinct robot embodiments is 22, creating a morphology bottleneck. Viewing robots as autonomous articulated bodies, we explore whether generative models can produce novel embodiments with valid kinematics and meaningful morphology.

Our contributions:
\textbf{(1)} We introduce an action conditioned flow matching framework that learns a velocity field in point space to generate deformations under given joint commands. 
\textbf{(2)} We extend the framework with joint conditioning on action and shape prior, enabling a single model to cover a category of articulated shapes and generate unseen instances with kinematically valid deformations.
\textbf{(3)} We evaluate on everyday objects and robots, demonstrating controllable generation, interpolation across instances, and accurate kinematic behavior under novel action sequences.

%% file: sec/2_related.tex
\section{Related Work}
\label{sec:RelatedWork}

\begin{table}[t]
\centering
{\scriptsize
\setlength{\tabcolsep}{1pt} 
\begin{tabular}{lccccc}
\toprule
\multirow{2}{*}{\textbf{Method}} &
\multirow{2}{*}{\textbf{Model}} 

  & \textbf{Shape}
  & \textbf{Latent}
  & \textbf{Articulated} 
  & \textbf{Color} \\
  
  && \textbf{Generation} 
  & \textbf{Interpolation} 
  & \textbf{Deformation} 
  & \textbf{Support} \\
\midrule
VSM~\cite{chen2022fully}        
& DeepSDF   & \xmark & \xmark & \cmark & \xmark \\
FFKSM~\cite{hu2025teaching}      
& NeRF      & \xmark & \xmark & \cmark & \xmark \\
\midrule
l-GAN~\cite{achlioptas2018learning}     
& GAN       & \cmark & \cmark & \xmark & \xmark \\
PointFlow~\cite{yang2019pointflow}
& CNF       & \cmark & \cmark & \xmark & \xmark \\
PVD~\cite{zhou20213d}  
& Diffusion & \cmark & \xmark & \xmark & \xmark \\
Point-E~\cite{nichol2022point}
& Diffusion & \cmark & \xmark & \xmark & \cmark \\
PSF~\cite{wu2023fast}   
& Flow Matching        & \cmark & \xmark & \xmark & \xmark \\
\midrule
\textbf{ArticFlow} & Flow Matching      & \cmark & \cmark & \cmark & \cmark \\
\bottomrule
\end{tabular}
}
\caption{\textbf{Compare with previous methods.} Rows 1--2 are neural simulation methods. Rows 3--7 are point cloud generation models.}
\label{tab:related-work}
\vspace{-1em}
\end{table}

\subsection{Point cloud generation}
Early methods adapted GAN and VAE to unordered point sets, for example, l-GAN~\cite{achlioptas2018learning} and hierarchical set-structured Set-VAE~\cite{kim2021setvae}. Flow based approaches model continuous deformations in data space: PointFlow~\cite{yang2019pointflow} learns a continuous normalizing flow over point clouds (trained on ShapeNet) by coupling a latent prior with a point-space flow.
Diffusion models followed, including point-voxel or pure point formulations such as PVD~\cite{zhou20213d}, PSF~\cite{wu2023fast}, and DiT-3D~\cite{mo2023dit}, which leverage score driven denoising with set-aware backbones. Latent diffusion has also been explored: LION~\cite{vahdat2022lion} trains an autoencoder for point sets and performs diffusion in the learned latent. Beyond unconditional generation, Point-E~\cite{nichol2022point} conditions on 2D image priors or text to produce colored point clouds, while the autoregressive tokenization of point sets (e.g., PointGPT~\cite{chen2023pointgpt}) treats generation as next-token prediction with permutation-robust tokenizers.
These methods improve fidelity and diversity while addressing set invariance, geometric realism, and efficient editing and inpainting. 

As generative models have grown in capability, architectural design has evolved from two-stage conditional generation, as in l-GAN and PointFlow, to direct one-stage formulations, such as the DDPM-based PVD and the rectified flow-matching framework of PSF.

\subsection{Articulated object modeling and generation}
Articulated objects are central to everyday tools and robotic manipulation, and interest is growing in both vision and robotics. Most prior work targets modeling and analysis: recovering parts, joints, and motion from images or 3D observations~\cite{liu2023paris,mu2021sdf,wang2019shape2motion,liu2023building}. Examples include MultiBodySync~\cite{huang2021multibodysync}, which segments rigid parts from point cloud sequences, and interaction-driven methods such as Ditto~\cite{jiang2022ditto} and Structure-from-Action~\cite{nie2022structure} that infer articulation parameters from robot interactions. Recent efforts translate perception into executable robot descriptions (Real2Code, URDFormer)~\cite{mandi2024real2code,chen2024urdformer}, and extend to complex robot structures (Watch\textminus It\textminus Move, AutoURDF)~\cite{noguchi2022watch,lin2025autourdf}.

Generative approaches are only beginning to be explored.
Image-conditioned pipelines synthesize articulated geometry from a single view~\cite{liu2024singapo,chen2025freeart3d}; mesh based methods deform or animate meshes with explicit part structures and joints~\cite{gao2025meshart,qiu2025articulate}; and graph based frameworks generate kinematic graphs before geometry~\cite{liu2024cage,lei2023nap}. These methods are often influenced by visual priors and guided by geometric assumptions (e.g., mesh connectivity), which may limit their flexibility in representing diverse or free-form geometries.
They often struggle to interpolate smoothly across instances and generate natural curved surfaces. 
By contrast, our method is self-contained and trained end-to-end; free form point cloud generation further enables smooth shape interpolations. 
Another related direction is \emph{neural robot simulation}~\cite{chen2022fully,hu2025teaching}, which learns task agnostic kinematic models from deformation samples in a differentiable form. 
However, prior works typically validate on one or a few specific robots, whereas our model captures a category of robots and generalizes to novel morphologies.

As task complexity rises for articulated mechanism generation, we revisit and adopt the two-stage design with flow matching to disentangle \emph{shape} and \emph{action}. 
Conditioned only on actions, our method functions as a neural simulator, predicting kinematically valid deformations of a given instance; 
conditioned only on a shape prior, it naturally generates diverse appearances while preserving the underlying kinematic functionality.

%% file: sec/3_method.tex
\section{Method}
\label{sec:Method}

\subsection{Notation and Overview.}
3D point-cloud generation aims to learn the sampling of permutation-invariant point sets from the data distribution. Methods based on diffusion and flow matching learn a mapping from a prior distribution \(P_{0}\), typically Gaussian noise \(\mathcal{N}(0,I)\), to the data distribution, denoted as \(P_{\text{data}}\). In this paper, we study action-conditioned flow matching that transports samples from a Gaussian prior to the shapes of articulated rigid bodies.

Given a dataset consisting of \(K\) sample pairs: point clouds \(\mathcal{X}=\{X^{i}\in\mathbb{R}^{N\times 3}\}_{i=1}^K\) and action vectors \(\mathcal{A}=\{A^{i}\in\mathbb{R}^{J}\}_{i=1}^K\), we jointly optimize two flow models.
The point flow transports data in point-cloud space from the Gaussian prior \(p_{0}=\mathcal{N}(0,I_{N\times 3})\) to the target set \(\mathcal{X}\).
The latent flow transports vectors from the normal distribution \(q_{0}=\mathcal{N}(0,I_{D})\) to the encoded latent-vector space \(\mathcal{Z}=\{Z^{i}\in\mathbb{R}^{D}\}_{i=1}^K\).
Here \(N\) is the number of points, \(J\) is the number of joints, and \(D\) is the dimension of the latent vector.
We denote the point flow as \(u_{\theta}(X_{t},t\,|\,Z_{a},Z_{x})\) and the latent flow as \(v_{\psi}(y_{t},t\,|\,Z_{a})\), where the encoded action latent is \(Z_{a}=E_{a}(A)\) and the encoded point-cloud latent is \(Z_{x}=E_{x}(X)\). 
For clarity of notation, we denote the pairwise (conditional) target velocities as $u_t$ and $v_t$, the corresponding marginal (expected) ground-truth velocities as $u^\star$ and $v^\star$, and the learnable velocity networks as $u_\theta$ and $v_\psi$.

In summary, the point flow learns the velocity field of the point cloud under the action and shape-prior conditions, and the latent flow learns to generate the shape-prior vector. We use point clouds \(\mathcal{X}\) and action vectors \(\mathcal{A}\) to train these two flow models. The goal is to train both flow models so that we can generate novel point clouds \(\hat{X}\) that follow the action condition \(\hat{A}\) with kinematically consistent deformations.

\begin{figure}[t!]
    \includegraphics[width=\columnwidth]{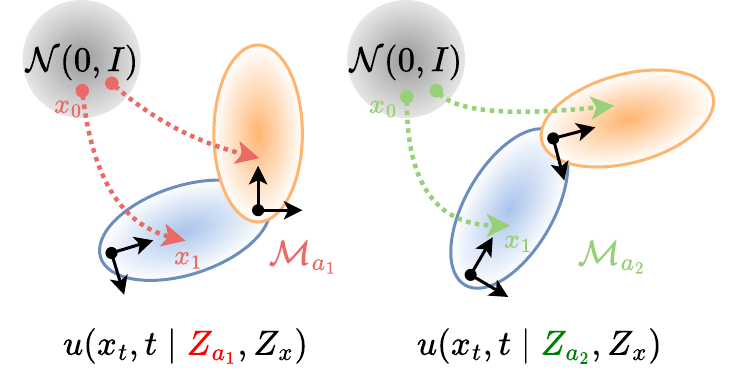}
    \caption{
    \textbf{Action-conditioned velocity fields.} 
    With the shape prior $Z_x$ fixed, changing the action embedding from $Z_{a_1}$ (left) to $Z_{a_2}$ (right) modulates the velocity fields $v(x,t \mid Z_a, Z_x)$. Each field transports states from the Gaussian prior ($x_0\sim\mathcal{N}(0, I)$) to an action-specific manifold $\mathcal{M}_{a}$ along the path $x_t$, producing distinct kinematic deformations. In our case, the target manifold consists of articulated rigid bodies represented as point clouds.
    }
    \label{fig:2-concept}
\end{figure}
\begin{figure*}[t!]
    \includegraphics[width=\textwidth]{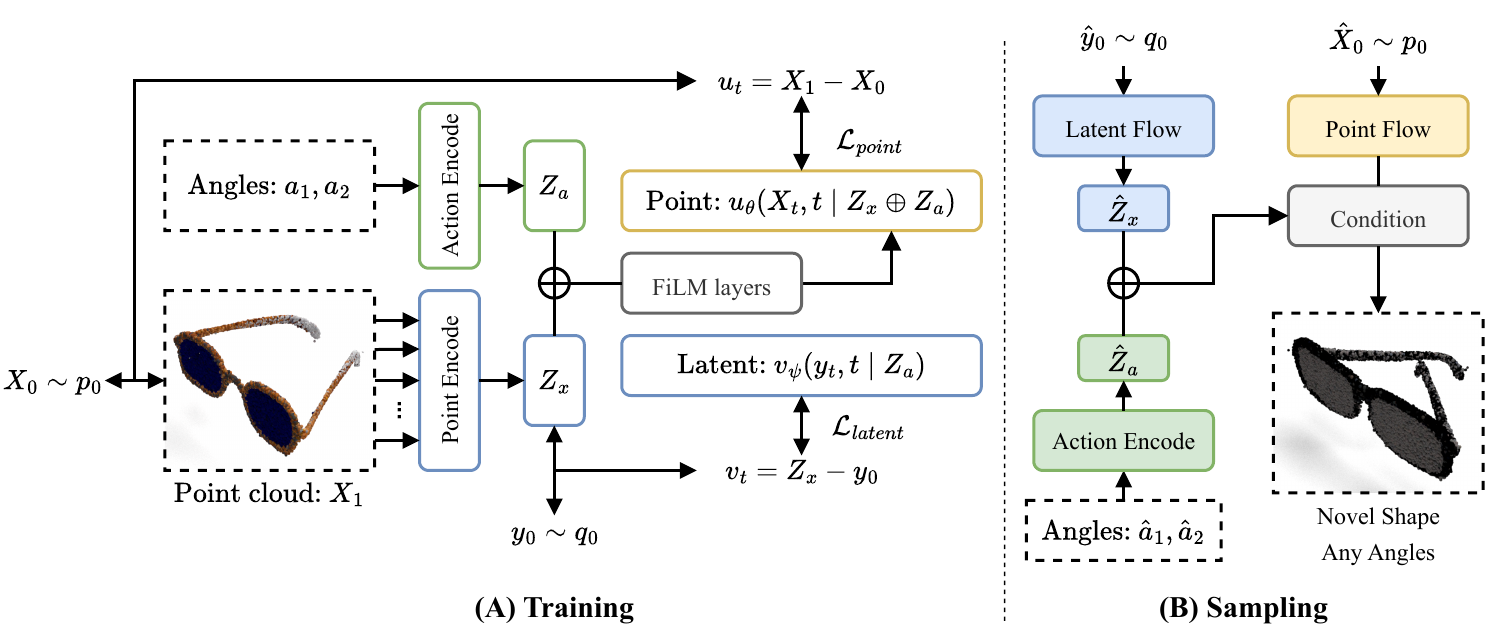}
    \caption{
    \textbf{Two-stage conditioned flow matching.} \textbf{(A).} For each training pair $(X_1, A)$, we encode the point cloud with PointNet~\cite{qi2017pointnet} to obtain the shape latent code $Z_x$ and encode the joint angles (Fourier layers~\cite{tancik2020fourier} + MLP) to obtain $Z_a$. We jointly optimize two flow-matching models: The \emph{point flow} $u_\theta$ predicts a conditional velocity in point space that transports points from a prior $p_0$ to the target set $X_1$ using straight-line pairings (target point set velocity $X_1-X_0$).
    The concatenated condition $(Z_x \oplus Z_a)$, added with time embedding, is injected using FiLM~\cite{perez2018film} layers. The \emph{latent flow} $v_\psi$ transports latent noise $y_0$ to the shape code $Z_x$, with target velocity $Z_x-y_0$. \textbf{(B).} At sampling time, We first draw latent noise $\hat y_{0}\sim q_{0}$ and an action set $(\hat a_1, \hat a_2)$. Integrating the latent ODE $\dot y_{t}=v_\psi$ forward from $t{=}0$ to $t{=}1$ yields a sampled shape latent code $\hat Z_{x}$. We then form the condition with $\hat Z_{x}$ and $\hat Z_{a}$, draw a point prior $\hat X_{0}\sim p_{0}$, and integrate the point ODE $\dot X_{t}=u_\theta$ to obtain the generated point cloud $\hat{X}$. This produces novel shapes with controllable articulation under arbitrary actions.
    }
    \label{fig:3-method}
\end{figure*}

\subsection{Action conditioned flow}
Flow Matching derives from the framework of Neural Ordinary Differential Equations (Neural ODEs)~\cite{chen2018neural}, 
where a data sample is obtained by integrating a continuous-time velocity field. 
Instead of directly fitting the ODE trajectory to data, Flow Matching learns the instantaneous velocity field that transports samples 
from a prior distribution to the data distribution. 
Intuitively, during training, each data pair $(X_0,X_1)$ defines a straight-line path $X_t=(1-t)X_0+tX_1$ 
and an associated velocity $X_1-X_0$. 
By minimizing the discrepancy between predicted and target velocities over random time steps and pairings, 
the network gradually converges to the \emph{marginal velocity field}, 
which represents the expected instantaneous transport direction at any intermediate state.
This marginal field defines a continuous flow that can be integrated as an ODE to generate samples from the learned distribution.

We now instantiate Flow Matching in point space by conditioning the velocity field on the joint action.
Given the encoded action latent $Z_a = E_a(A)$, the point flow aims to learn a velocity field
$u_\theta(X_t, t \mid Z_a)$ that transports samples from the Gaussian prior to the articulated shapes in data space.
As the given $X_t$ may correspond to multiple $(X_0, X_1)$ pairs, we learn the \emph{marginal velocity:}
\begin{equation}
u^\star(X, t \mid Z_a)
=
\mathbb{E}\!\left[\,u_t \;\middle|\; X_t = X,\, t,\, Z_a \right]
\label{eq:marginal-velocity}
\end{equation}
which represents the expected instantaneous velocity of a point set conditioned on action.
The \emph{conditional velocity} $u_t = X_1-X_0$.
The flow matching loss minimizes the squared error between the network prediction and the target velocity:
\begin{equation}
\mathcal{L}_{\text{point}}
=
\mathbb{E}_{t, X_0, X_1}
\!\left[
\big\|
u_\theta(X_t, t \mid Z_a)
-
u_t
\big\|_2^2
\right].
\label{eq:point-loss}
\end{equation}
Here $t$ is sampled from a time distribution (e.g., $\mathrm{Beta}(2,1)$) and $X_t = (1-t)X_0 + tX_1$ is the interpolated point set.
After training, new samples are generated by solving the ODE
\begin{equation}
    \dot{X}_t = u_\theta(X_t, t \mid Z_a), \quad \hat X_0 \sim p_0
\label{eq:point-ode}
\end{equation}
which evolves the initial noise $\hat X_0$ toward a valid articulated shape $\hat X$ under the action code $Z_a$.
In practice, we apply numerical integration, such as Euler updates
$X_{k+1} = X_k + h\,u_\theta(X_k, t_k \mid Z_a)$, where $h$ denotes the step size.

\subsection{Action and shape-prior conditioned flow}
In the two-stage conditioned flow matching, as shown in Figure~\ref{fig:3-method} (A), we jointly optimize two flow matching models.
Equation~\ref{eq:marginal-velocity} will be extended with two latent conditions:
\begin{equation}
u^\star(X, t \mid Z_a, Z_x)
=
\mathbb{E}\!\left[\,u_t \;\middle|\; X_t = X,\, t,\, Z_a, Z_x \right]
\label{eq:marginal-velocity-x}
\end{equation}
Here $Z_x$ is the encoded shape latent vector, $E_x(X)$. We then introduce the second flow matching model $v_\psi$, along with the corresponding \emph{marginal velocity} in latent space:
\begin{equation}
v^\star(y, t \mid Z_a)
=
\mathbb{E}\!\left[\,v_t \;\middle|\; y_t = y,\, t,\, Z_a \right]
\label{eq:marginal-velocity-y}
\end{equation}
We use straight-line pairings to define the conditional velocities. 
In point space, we sample $X_0 \!\sim\! p_0$ and $X_1 \!\sim\! p_{\text{data}}(\cdot \mid Z_a, Z_x)$, 
construct the interpolated point set $X_t = (1 - t) X_0 + t X_1$, 
and define the pairwise velocity $u_t = X_1 - X_0$. 
Similarly, in latent space, we sample $y_0 \!\sim\! q_0$, 
set $y_t = (1 - t) y_0 + t Z_x$, 
and define $v_t = Z_x - y_0$. 
As shown in algorithm\ref{alg:articflow-training}, we use a time sampling distribution, typically $\mathrm{Beta}(2,1)$, which places higher density near $t=1$ and improves training stability in the later steps of the flow.

In our implementation, the shape latent $Z_x$ and the action latent $Z_a$ are designed to have the same dimensionality as time embedding, so we can combine them by elementwise addition, which is then passed through FiLM layers\cite{perez2018film}.
The FiLM module generates feature-wise scale and shift parameters, allowing the condition to modulate intermediate activations of the velocity networks through affine transformations. 
This provides an efficient and expressive way to inject both shape and action information into the flow-matching models without altering their core architectures.

Generally, this two stage design factorizes shape and action into two independent latent spaces. 
The latent flow $v_\psi$ learns a distribution over shape codes by transporting noise toward the encoded vectors $Z_x$, effectively compressing geometric variability into a compact latent prior. 
Conditioned on a sampled shape code, the point flow $u_\theta$ focuses solely on action driven deformation, learning how each instance moves under different joint commands. 
This separation allows us to first generate a plausible shape latent that matches the data distribution and then apply the action conditioned point flow to produce articulated motions with disentangled control over shape and action.

\subsection{Sampling and Interpolation}
At sampling time (Figure~\ref{fig:3-method} (B) and Algorithm~\ref{alg:articflow-sampling}), we first command an input action $\hat{A}$ and encode it into a latent vector $\hat{Z}_a = E_a(\hat{A})$. 
We then sample latent noise and integrate the latent flow ODE to obtain a shape code:
\begin{equation}
\dot{y}_t = v_\psi(y_t, t \mid \hat{Z}_a)
\end{equation}
with $y_0 \sim q_0, \; \hat{Z}_x = y_1$.
Given this integrated shape latent $\hat{Z}_x$ and the action latent $\hat{Z}_a$, we initialize a point cloud and integrate the point flow ODE:
\begin{equation}
\dot{X}_t = u_\theta(X_t, t \mid \hat{Z}_x, \hat{Z}_a)
\end{equation}
with $X_0 \sim p_0, \; \hat{X} = X_1$.
This equation yields a novel articulated shape $\hat{X}$ consistent with the conditioned shape and action. 
In practice, both ODEs are solved numerically using Heun's method (improved Euler) with $S_{\text{lat}}$ and $S_{\text{pt}}$ integration steps for the latent and point flows, respectively. The detailed calculation steps are specified in the algorithm block.
In general, the sampling framework takes only the joint conditions as input. Given the trained flow models $v_\psi$ and $u_\theta$, it generates new objects whose articulation follows the commanded joint configurations.

For interpolation, we simply perform spherical linear interpolation (slerp) in
the shape latent space. Given two sampled shape latents
$\hat Z_x^{(0)}$ and $\hat Z_x^{(1)}$, we generate intermediate latent codes
$\hat Z_x(\alpha)$ with $\alpha \in [0,1]$ by
\[
\operatorname{slerp}\big(\hat{Z}_x^{(0)},\hat{Z}_x^{(1)}; \alpha\big)= \frac{\sin((1-\alpha)\phi)}{\sin\phi}\,\hat{Z}_x^{(0)} + \frac{\sin(\alpha\phi)}{\sin\phi}\,\hat{Z}_x^{(1)}
\]
where $\phi$ represents the angle between the two latent vectors $\hat{Z}_x^{(0)}$ and $\hat{Z}_x^{(1)}$, which is the geodesic distance on the unit hypersphere connecting their normalized directions.

Unlike static point cloud generation work on large datasets such as ShapeNet~\cite{chang2015shapenet}, 
PartNet-Mobility~\cite{xiang2020sapien} provides limited instances per category. 
Although we wish for the latent shape to capture kinematics-irrelevant features, learning a meaningful latent space with limited samples is challenging. We therefore validate two variants. 
First, we condition the shape latent flow on both the category shape and the 
action deformation by injecting the action latent $\hat Z_a$. The same as illustrated above. 
In the second version, we remove the action condition and train an unconditional shape latent flow with large shape diversity.
In the later stage of training, we additionally attach a conditional adversarial module with a gradient reversal layer (GRL)~\cite{ganin2016domain} to remove action information 
from the shape latent. 
We compare these variants, denoted as ArticFlow (Cond-latent) and ArticFlow (Adv-latent), in the Experiments~\ref{sec:Experiments} section.

ArticFlow also supports colored point cloud generation by simply concatenating 3D positions and RGB values. Specifically, we extend the point flow velocity field to operate on 6D vectors and train it with input $X_0$ and target $X_1$, both of which contain position and color channels.

\begin{algorithm}[t]
\caption{\textbf{ArticFlow Training (Flow Matching)}}
\label{alg:articflow-training}
\begin{algorithmic}[1]
\Require Dataset $\{(X^{i},A^{i})\}_{i=1}^K$
\Statex priors $p_0=\mathcal N(0,I_{N\times3})$, $q_0=\mathcal N(0,I_D)$
\Statex velocity fields $u_\theta, v_\psi$, optimizer $\mathrm{Opt}$.
\For{step $=1$ to $T$}
    \State Sample minibatch $X_1, A$
    \State $Z_x \gets E_x(X_1)$,\quad $Z_a \gets E_a(A)$ \Comment{Encode}
    \State Sample $t_x \sim \mathrm{Beta}(2,1)$,\quad$X_0 \sim p_0$ 
    \Comment{Point flow}
    \State $X_t \gets (1-t_x)\,X_0 + t_x\,X_1$
    \State $u_t \gets X_1 - X_0$,\quad $u \gets u_\theta(X_t, t \mid Z_x, Z_a)$
    \State $\mathcal{L}_{\text{point}} \gets \mathrm{MSE} (u, u_t)$
    \State Sample $t_z \sim \mathrm{Beta}(2,1)$,\quad$y_0 \sim q_0$ 
    \Comment{Latent flow}
    \State $y_t \gets (1-t_z)\,y_0 + t_z\,Z_x$
    \State $v_t \gets Z_x - y_0$,\quad $v \gets v_\psi(y_t, t \mid Z_a)$
    \State $\mathcal{L}_{\text{latent}} \gets \mathrm{MSE} (v, v_t)$
    \State $\mathcal{L} \gets \mathcal{L}_{\text{point}} + \lambda\,\mathcal{L}_{\text{latent}}$
    \State $\mathrm{Opt}.\text{step}()$ \Comment{Optimize $E_x, E_a, u_\theta, v_\psi$}
\EndFor
\end{algorithmic}
\end{algorithm}

\begin{algorithm}[t]
\caption{\textbf{ArticFlow Sampling (Heun RK2)}}
\label{alg:articflow-sampling}
\begin{algorithmic}[1]
\Require Trained $v_\psi, u_\theta$, encoder $E_a$, action $\hat A$
\Statex priors $q_0=\mathcal N(0,I_D)$, $p_0=\mathcal N(0,I_{N\times3})$ 
\Statex steps $S_{\mathrm{lat}},S_{\mathrm{pt}}$

\Procedure{Heun}{field, $x_0$, cond, $S$} \Comment{Improved Euler}
  \State $h\gets 1/S$;\; $x\gets x_0$;\; $t\gets 0$
  \For{$s=1$ \textbf{to} $S$}
    \State $k_1 \gets \text{field}(x, t \mid \text{cond})$
    \State $\bar x \gets x + h\,k_1$
    \State $k_2 \gets \text{field}(\bar x, t{+}h \mid \text{cond})$
    \State $x \gets x + \tfrac{h}{2}(k_1{+}k_2)$;\; $t \gets t + h$
  \EndFor
  \State \textbf{return} $x$
\EndProcedure

\State $\hat Z_a \gets E_a(\hat A)$
\State $\hat Z_x \gets \Call{Heun}{v_\psi,\; y_0\!\sim\!q_0,\; \hat Z_a,\; S_{\mathrm{lat}}}$
\State $\hat X \gets \Call{Heun}{u_\theta,\; X_0\!\sim\!p_0,\; \hat Z_x,\hat Z_a,\; S_{\mathrm{pt}}}$
\State \textbf{return} $\hat X$
\end{algorithmic}
\end{algorithm}

%% file: sec/4_experiments.tex
\begin{figure*}[t!]
    \includegraphics[width=\textwidth]{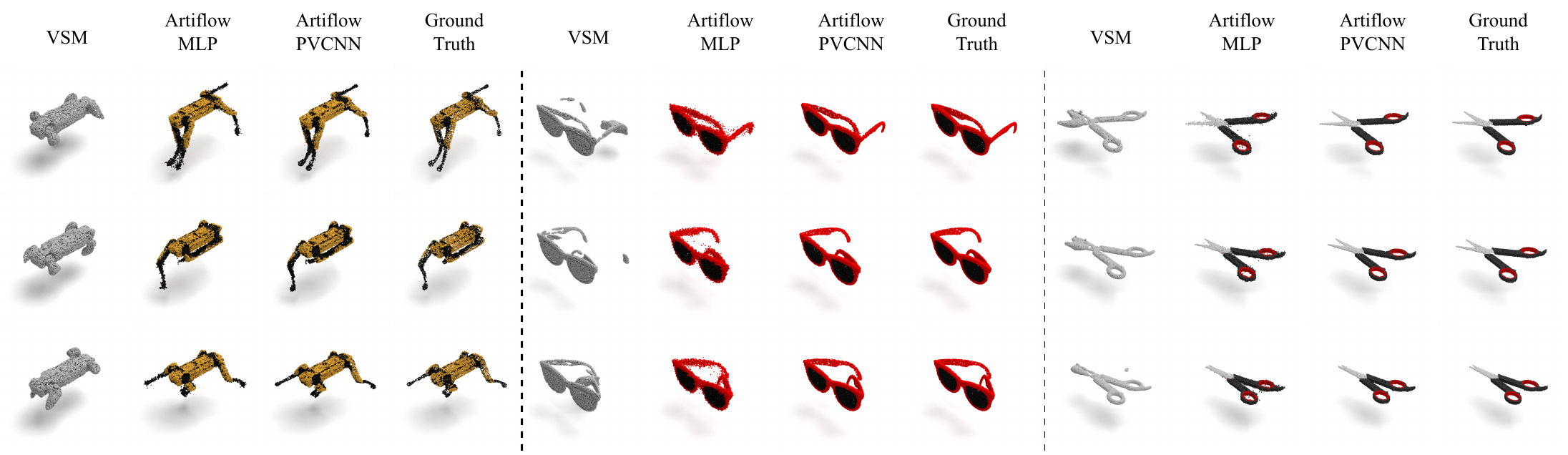}
    \caption{\textbf{Action-conditioned flow: qualitative results.}
    For a fixed object and varying joint actions, we compare the deformed point clouds predicted by VSM, ArticFlow-MLP, ArticFlow-PVCNN, and the ground truth. Conditioned only on the action latent, ArticFlow produces shapes whose articulation more closely matches the target kinematics than VSM.}
    \label{fig:4-sim}
\end{figure*}

\begin{figure*}[t!]
    \includegraphics[width=\textwidth]{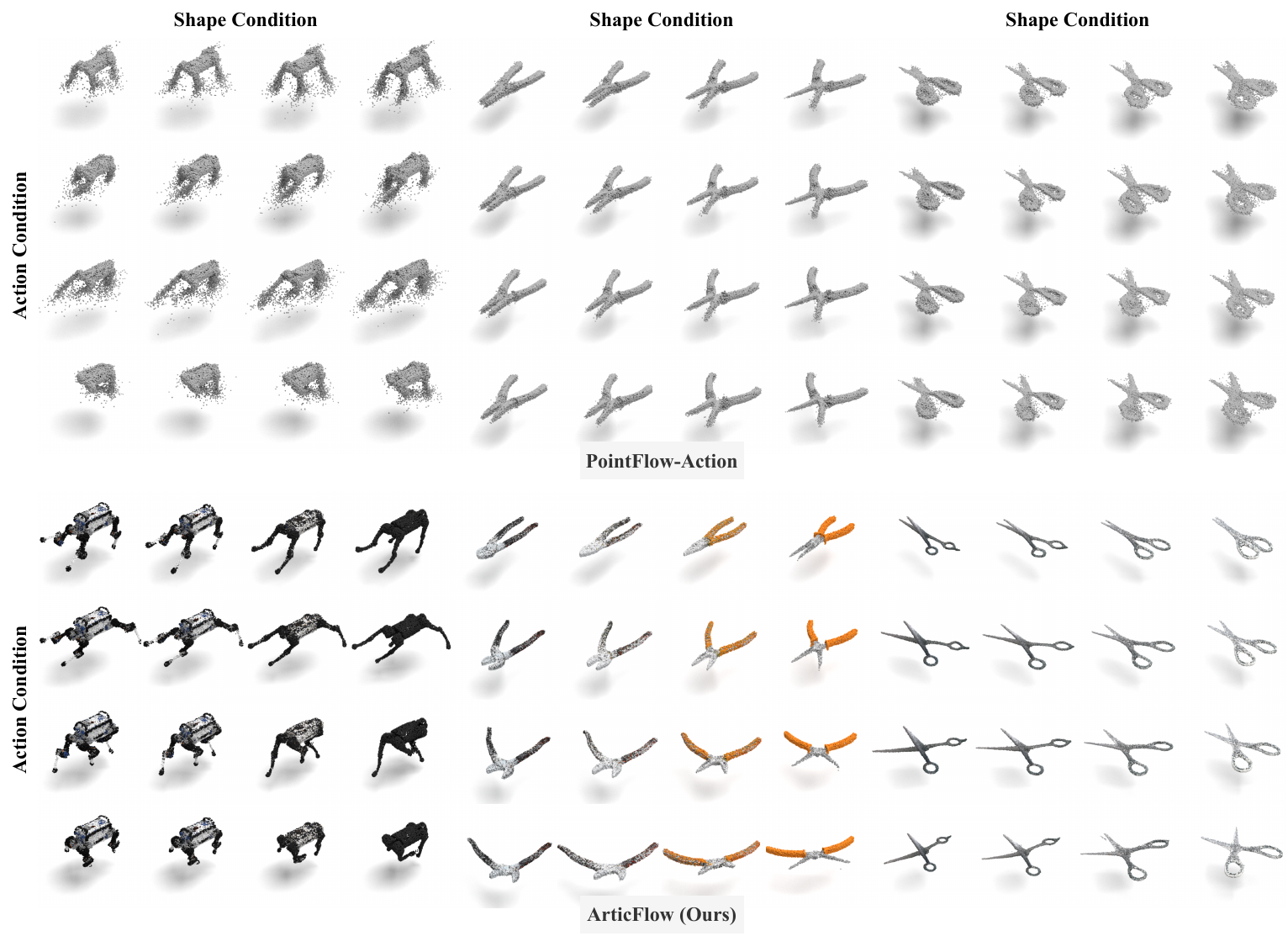}
\caption{
\textbf{Action and shape-prior conditioned flow, qualitative results.}
When conditioned on both the shape latent and the action latent, we compare
our method (ArticFlow) with the action-conditioned PointFlow baseline.
Columns vary the shape prior (different object instances) and rows sweep the
action condition. ArticFlow produces higher-quality surfaces and kinematically valid deformations. For the PointFlow-Action baseline, simple 1-DoF objects show only minor deformation across angle conditions, and more complex kinematics lose fine shape details, such as the robot legs.
}
    \label{fig:5-gen}
\end{figure*}

\begin{figure*}[t!]
    \includegraphics[width=\textwidth]{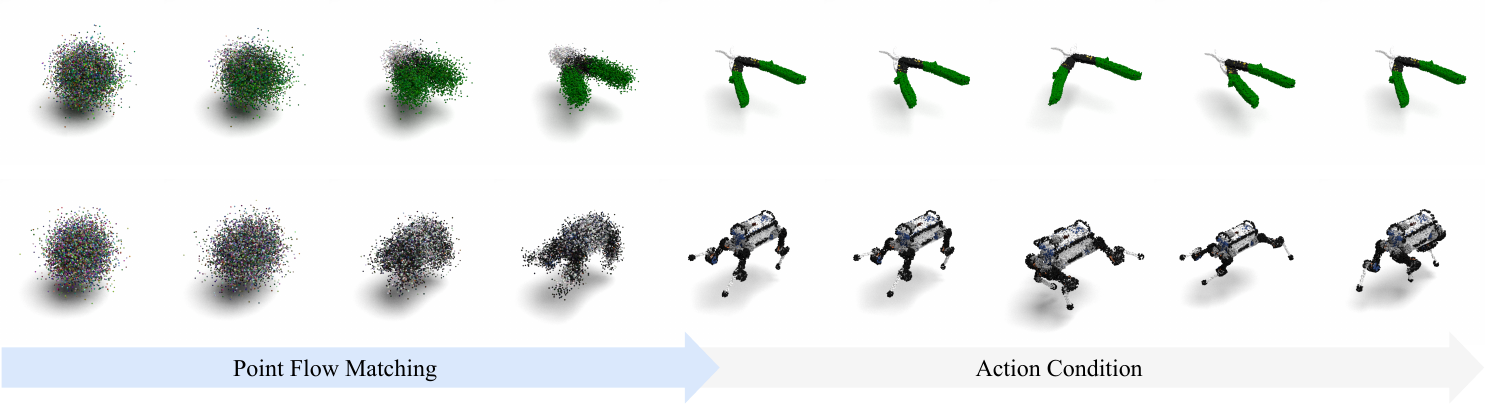}
    \caption{
    \textbf{Generation process and action conditioning.}
    Two examples from models trained on pliers (top) and quadruped robots (bottom).
    On the left, point flow matching transforms Gaussian noise into object
    shape; on the right, sampling random joint commands produces diverse
    articulated configurations of the same shape. Shape and color are generated jointly from 6D Gaussian prior.
    }
    \label{fig:6-demo}
\end{figure*}

\section{Experiments}
\label{sec:Experiments}
\subsection{Dataset and Metrics}
We validate our method on five categories: pliers, scissors, and eyeglasses
from PartNet-Mobility, robot arms and quadrupeds from MuJoCo Menagerie.
For each PartNet-Mobility object, we use 50 kinematic samples, and for each
robot we use 1000 action configurations, with a $5{:}1$ train–test split.
For PartNet categories, each point cloud contains 20k points for flow
training. For the robot categories, we use fewer points per sample, namely
4{,}096, since there are many more kinematic samples per object. We adopt
these settings for all quantitative and qualitative experiments. During
sampling and evaluation, we resample each prediction to 20k points so that
all evaluation metrics are computed on point clouds of the same size. 
Within each category, the number of degrees of freedom (DoF) can vary across
instances; therefore, we pad the action vectors with zeros up to the maximum DoF in
that category. For complex structures such as robots, the rotation
directions along the kinematic tree may differ between models; before
training, we normalize all joint rotations by aligning them to the Denavit–Hartenberg (DH) convention for the robot dataset.

The goal of our quantitative experiments is to evaluate whether ArticFlow can
produce kinematically correct shapes and how accurately it follows the commanded
joint actions. Direct evaluation of novel articulated shapes is difficult,
since ground-truth deformations for unseen objects are not available.
Therefore, we measure kinematic accuracy on latent codes decoded from
articulated objects in the dataset, conditioning on unseen action configurations. We compare ArticFlow with the action-conditioned PointFlow baseline. We also consider a single-object setting, where the model is trained on one specific robot and conditioned
only on the action, and compare it with the visual self-modeling~\cite{chen2022fully} (VSM)
method designed for robot representation learning.

For evaluation, we use both the L2-Chamfer Distance (CD) and the Earth Mover’s Distance (EMD) between the predicted and ground-truth point clouds. Chamfer captures overall shape similarity and is efficient to compute, while EMD provides a transport based measure that is more sensitive to local misalignment. Reporting both metrics provides a more complete assessment of geometric accuracy.

\input{sec/table2}
\input{sec/table3}
\subsection{Qualitative Results}
For baseline comparison, we adapt the original PointFlow~\cite{yang2019pointflow} model into an action-conditioned generator by injecting the action information through FiLM layers while keeping the CNF-based training framework, denoted as PointFlow-Action. To compare fairly with VSM, which requires high-quality surface normals, we additionally implement its training pipeline and construct a dedicated dataset: starting from the same point clouds used by ArticFlow, we reconstruct watertight meshes and compute normals before training VSM.
Figure~\ref{fig:4-sim} shows qualitative results for the
action conditioned setting. For a fixed object and varying joint actions,
ArticFlow produces deformations that more closely follow the target
kinematics than VSM, especially in thin structures such as eyeglass legs and
robot limbs. Between our two backbones, the PVCNN~\cite{liu2019point} variant yields sharper and
more complete surfaces than the per-point MLP. 
Figure~\ref{fig:5-gen} presents qualitative results for
joint shape and action conditioned generation. Compared with the
action conditioned PointFlow baseline, ArticFlow generates higher-quality
surfaces and more kinematically consistent deformations across different
instances within each category. In addition, ArticFlow
naturally supports color generation, whereas VSM and PointFlow do not capture RGB features.

\subsection{Quantitative Results}

In the action condition setting (Table~\ref{tab:artiflow_sim}), ArticFlow substantially
outperforms VSM across categories in CD and EMD, indicating more accurate kinematic deformation for appearance fixed objects. Comparing backbones,
the PVCNN variant consistently improves upon the per-point MLP architecture, especially in EMD, suggesting that the point-convolution module captures local geometric structure more effectively and therefore reduces the transport cost between
the predicted and ground-truth point sets.

In the two stage generation setting (Table~\ref{tab:artiflow_gen}), all ArticFlow variants
outperform the action-conditioned PointFlow baseline, showing that our joint
shape–action modeling yields more accurate articulated shapes. The different
latent-flow designs achieve similar CD and EMD, with the unconditional latent
and the adversarially regularized latent performing slightly better overall,
indicating that discouraging kinematic information in the shape latent can
improve generalization.

\subsection{Applications and Future Work}

\begin{figure}[t!]
    \includegraphics[width=\columnwidth]{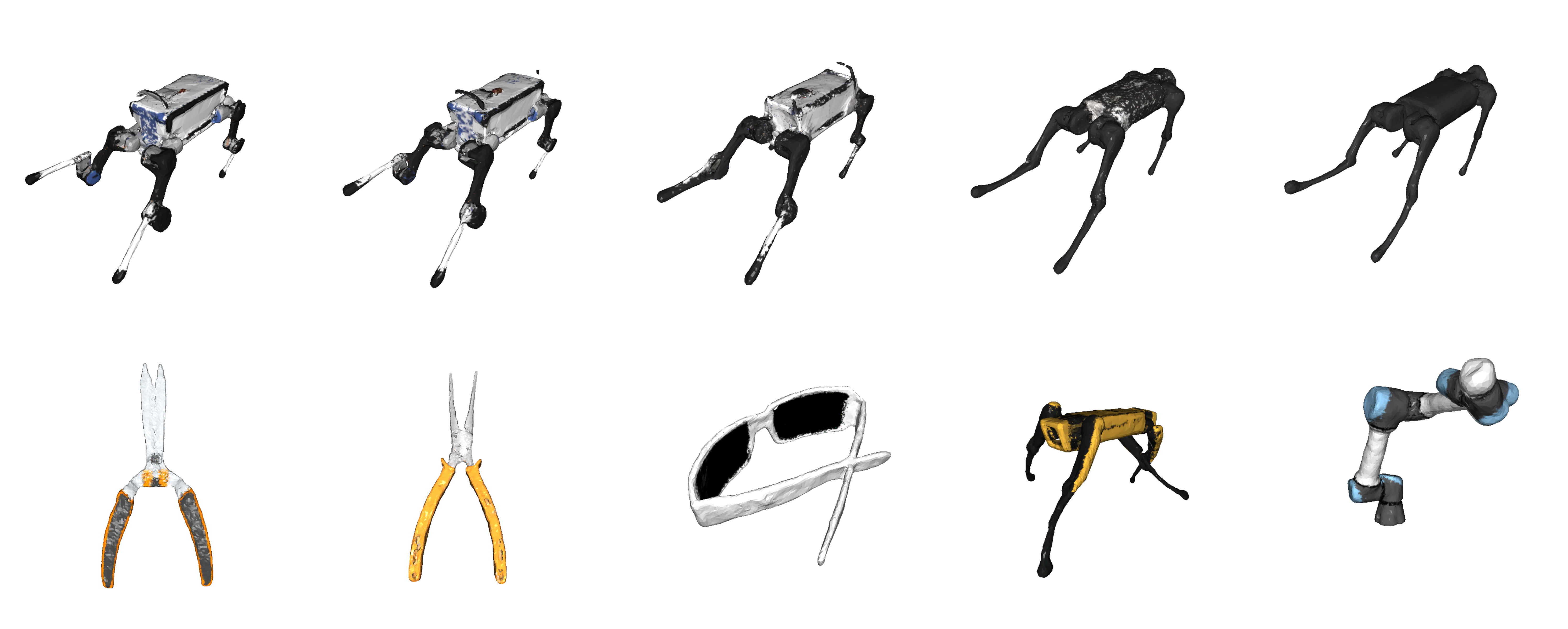}
    \caption{
    \textbf{Point cloud to mesh.}
    We demonstrate interpolated point clouds from ArticFlow and their conversion to
    watertight meshes. The top row shows an interpolation sequence between two
    legged robots. The bottom row presents random samples from other
    categories. We preserve the color from point cloud to mesh.
    }
    \label{fig:7-mesh}
\end{figure}

Figures~\ref{fig:6-demo} and~\ref{fig:7-mesh} illustrate that
ArticFlow supports both generation and simulation oriented reconstruction.
Figure~\ref{fig:6-demo} shows how our point-flow first maps 6D Gaussian
noise to a canonical colored shape and then uses action conditioning to
produce diverse articulated configurations. Figure~\ref{fig:7-mesh}
demonstrates that these generated point clouds can be converted into
high-quality watertight meshes using recent point cloud to mesh work~\cite{yu2025pointdreamer}. Bridging with popular articulation modeling
methods, ArticFlow outputs could be further transformed
into simulation-ready description files (e.g., URDF), closing the loop
towards fully generated, simulation-ready articulated models.

%% file: sec/table2.tex
\begin{table*}[t]
\centering
{\small
\setlength{\tabcolsep}{2pt}  
\begin{tabular}{l|cc|ccc|cc|ccc}
\toprule
\multirow{2}{*}{\textbf{Simulation}} &
\multicolumn{5}{c|}{\textbf{CD} $\downarrow$} &
\multicolumn{5}{c}{\textbf{EMD} $\downarrow$} \\
\cmidrule(lr){2-6}
\cmidrule(lr){7-11}
& \textbf{Arms} & \textbf{Quadrupeds} & \textbf{Pliers} & \textbf{Scissors} & \textbf{Eyeglasses} &
\textbf{Arms} & \textbf{Quadrupeds} & \textbf{Pliers} & \textbf{Scissors} & \textbf{Eyeglasses} \\
\midrule
VSM~\cite{chen2022fully} & 12.4763 & 14.5676 & \textbf{0.0527} & 1.7208 & 10.7652 & 57.7256 & 66.9668 & \textbf{8.7149} & 45.9159 & 76.1755 \\

ArticFlow(MLP) & \textbf{0.0599} & 0.1853 & 0.1232 & 0.0780 & 0.1522 & 20.9800 & 15.9860 & 14.9094 & 14.6320 & 29.1480 \\

ArticFlow(PVCNN) & 0.6228 & \textbf{0.1675} & \textbf{0.1240} & \textbf{0.0481} & \textbf{0.1400} & \textbf{11.1780} & \textbf{5.8883} & \textbf{9.9892} & \textbf{8.5441} & \textbf{4.4339} \\
\bottomrule
\end{tabular}
}
\caption{
\textbf{Baseline comparison on action-conditioned simulation.}
Chamfer Distance (CD) and Earth Move Distance (EMD) values $\times 10^{3}$.
}
\label{tab:artiflow_sim}
\vspace{-0em}
\end{table*}

%% file: sec/table3.tex
\begin{table*}[t]
\centering
{\small
\setlength{\tabcolsep}{2pt}  
\begin{tabular}{l|cc|ccc|cc|ccc}
\toprule
\multirow{2}{*}{\textbf{Generation}} &
\multicolumn{5}{c|}{\textbf{CD} $\downarrow$} &
\multicolumn{5}{c}{\textbf{EMD} $\downarrow$} \\
\cmidrule(lr){2-6}
\cmidrule(lr){7-11}
& \textbf{Arms} & \textbf{Quadrupeds} & \textbf{Pliers} & \textbf{Scissors} & \textbf{Eyeglasses} &
\textbf{Arms} & \textbf{Quadrupeds} & \textbf{Pliers} & \textbf{Scissors} & \textbf{Eyeglasses} \\
\midrule
PointFlow-Action    
& 0.2824 & 1.4416 & 0.3591 & 0.4829 & 1.8308  
& 62.8441 & 79.5037 & 73.3096 & 76.0229 & 182.6917  \\
\text{ArticFlow(Uncond-latent)}
& 0.6061 & 0.2502 & 0.0281 & 0.0308 & 0.0207
& 16.8010 & 8.2525 & \textbf{4.4291} & \textbf{6.0803} & 4.3361  \\
\text{ArticFlow(Cond-latent)}
& 0.6776 & 0.2651 & \textbf{0.0281} & \textbf{0.0295} & 0.0366  
& 18.2628 & 8.7380 & 4.5394 & 6.2370 & 4.2984  \\
\text{ArticFlow(Adv-latent)}
& \textbf{0.0616} & \textbf{0.2194} & 0.0306 & 0.0361 & 0.0311 
& \textbf{9.8705} & \textbf{7.9411} & 4.6393 & 6.4390 & \textbf{3.6163}\\
\bottomrule
\end{tabular}
}
\caption{
\textbf{Baseline comparison on action and shape-prior conditioned generation.}
CD and EMD values $\times 10^{3}$.
}
\label{tab:artiflow_gen}
\vspace{-0em}
\end{table*}

%% file: sec/5_discussion.tex
\section{Discussion}
\label{sec:Discussion}
{\bf{Limitations.}} 
Our approach also has several limitations. First, generation is currently limited to the categories and articulation patterns present in the training datasets; extending to more diverse objects would require additional curated data. Second, joint directions and ranges must be aligned for high-quality training, which limits our ability to directly learn from real-world articulated data. Finally, while the model can be queried with actions outside the training range, such extrapolated poses may yield unrealistic geometries. 

\noindent{{\bf{Conclusion.}}}
This paper shows that action-conditioned flow matching provides an effective framework for controllable articulated 3D generation. ArticFlow combines a latent flow for shape priors with a point flow for action-conditioned deformations, allowing a single model to represent diverse articulated categories and generalize across actions. In both PartNet-Mobility and MuJoCo Menagerie experiments, ArticFlow outperforms the action-conditioned PointFlow and VSM in terms of deformation accuracy. We expect this framework to scale to broader object categories and to integrate with simulation-ready description formats in future work.

%% file: sec/X_suppl.tex
\twocolumn[{%
  \renewcommand{\twocolumn}[1][]{#1}
  \maketitlesupplementary
\begin{center}
  \includegraphics[width=\textwidth]{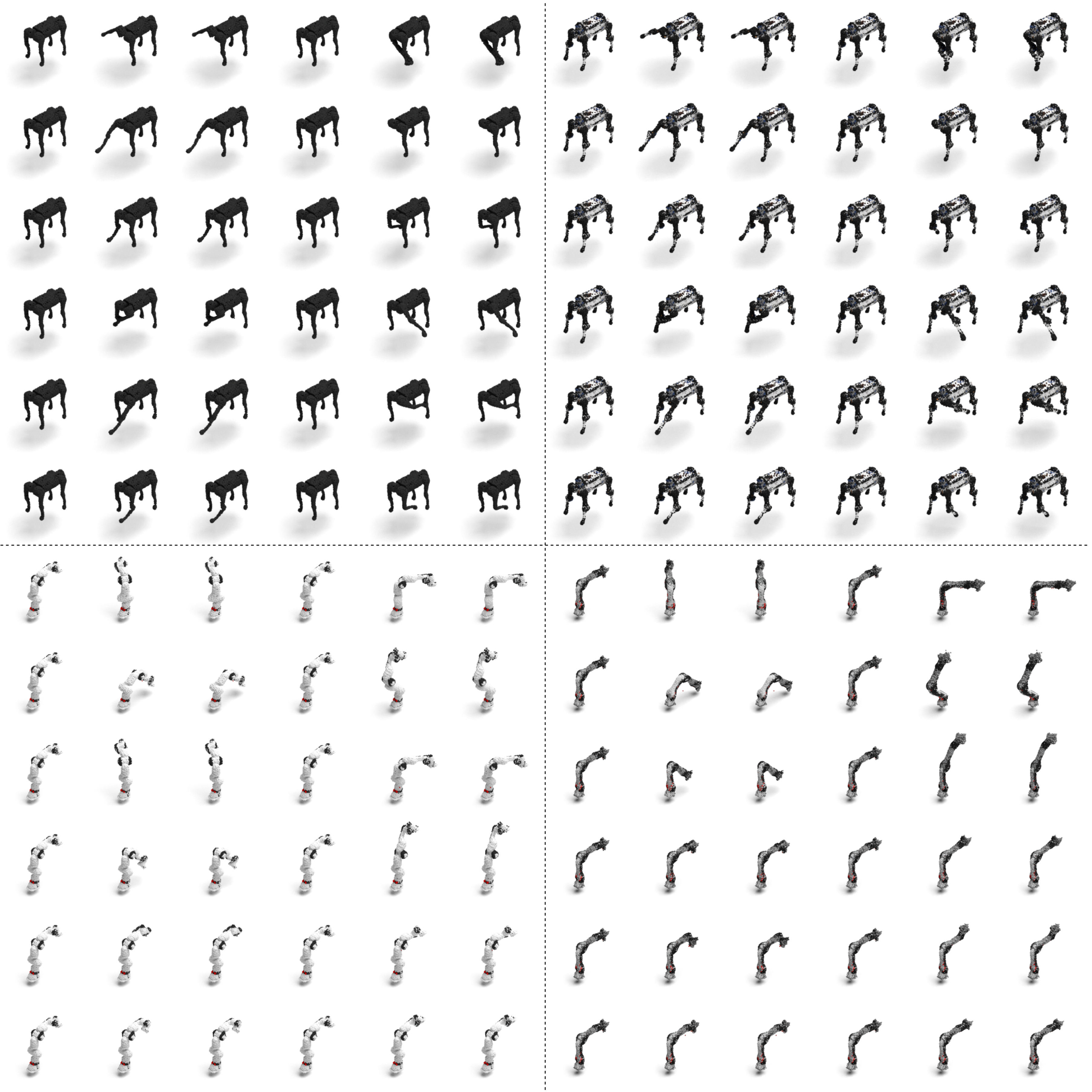}
  \captionsetup{type=figure, hypcap=false} 
  \caption{
   \textbf{Kinematic control validation.} 
Left: sampled quadruped robot and robotic arm from the dataset. 
Right: two interpolated novel robot shapes under action control. 
For visualization, we sweep the first six joints one by one.
    }
  \label{fig:teaser}
\end{center}
}]